\pdfoutput=1

\documentclass[11pt]{article}

\usepackage{acl}

\usepackage{times}
\usepackage{latexsym}

\usepackage[T1]{fontenc}

\usepackage{graphicx} 
\usepackage{subcaption} 

\usepackage[utf8]{inputenc}

\usepackage{microtype}
\usepackage{booktabs}
\usepackage{inconsolata}
\usepackage{multirow}
\usepackage{array} 
\usepackage{verbatim} 
\usepackage{listings} 
\usepackage[spacesep,definevectors]{easyvector}
\usepackage{amsmath}
\usepackage{mathtools}
\usepackage{xspace}
\usepackage{algpseudocode}
\usepackage{algorithm}

\newcommand{\fixme}[1]{\footnote{\textbf{FIXME!!!} #1}}
\newcommand{\OurMODEL}{MoRAL}
\newcommand{\eat}[1]{}
\newcommand{\asif}[1]{\textcolor{blue}{#1}}
\newcommand{\warn}[1]{\textcolor{red}{#1}}
\DeclareUnicodeCharacter{0301}{\hspace{-1ex}\'{ }}

\title{\OurMODEL{}: MoE Augmented LoRA for LLMs' Lifelong Learning}

\author{
Shu Yang$^{*,1,2,3}$,
Muhammad Asif Ali$^{*,1,2}$,
Cheng-Long Wang$^{1,2}$,
Lijie Hu$^{\dagger,1,2,4}$,
and Di Wang$^{\dagger,1,2,4}$ \\
$^1$Provable Responsible AI and Data Analytics (PRADA) Lab\\
$^2$King Abdullah University of Science and Technology\\
$^3$University of Macau \quad $^4$SDAIA-KAUST AI \\
$^*$Equal Contribution \quad $^\dagger$Corresponding Author
}

\begin{document}
\maketitle
\begin{abstract}
Adapting large language models (LLMs) to 
new domains/tasks and enabling them to be 
efficient lifelong learners is a pivotal challenge. 
In this paper, we propose \OurMODEL{}, i.e., 
\underline{\textbf{M}}ixture-\underline{\textbf{o}}f-Experts augmented Low 
\underline{\textbf{R}}ank \underline{\textbf{A}}daptation for \underline{\textbf{L}}ifelong 
learning. \OurMODEL{} combines the multi-tasking abilities 
of MoE with the fine-tuning abilities of LoRA 
for effective life-long learning of LLMs. 
In contrast to the conventional approaches that use 
factual triplets as inputs~\OurMODEL{} 
relies on simple question-answer pairs,
which is a more practical and effective strategy for robust and efficient learning.
Owing to new data settings, we introduce a new evaluation benchmark namely: Life Long Learning of LLM (5L-bench) encompassing a newly curated dataset of question-answer pairs, and a set of evaluation metrics for rigorous evaluation of \OurMODEL{} 
in open-book and closed-book settings.
Experimental evaluation shows
(i) LLMs learn fast in open-book settings with up to 30.15\% 
improvement in "RA" for Phi-2-2.7B compared to closed-book 
(for models fine-tuned with~\OurMODEL{});
(ii)~\OurMODEL{} shows higher performance improvement for models with a greater number of parameters; 
(iii)~\OurMODEL{} is robust to catastrophic forgetting offering better
knowledge retention compared to baselines.

\end{abstract}

\section{Introduction}

Large language models (LLMs) trained using massive 
computational clusters and expansive datasets, have 
demonstrated impressive proficiency in natural language processing~\cite{zhao2023survey,kaddour2023challenges}. 
These models excel in a variety of downstream tasks, such as machine translation~\cite{zhu2023multilingual, xu2024contrastive}, grammatical error correction~\cite{fang2023chatgpt, wu2023chatgpt} etc. The success of LLM arises from the powerful knowledge processing and compression capabilities~\cite{zhu2023survey, huang2023approximating, delétang2023language}, which allow LLMs to construct information 
in a way that is somehow similar to humans, and even complete never-before-seen tasks~\cite{grosse2023studying,kirk2024understanding}. 

\begin{figure}[t]
\centering
\includegraphics[width=0.5\textwidth]{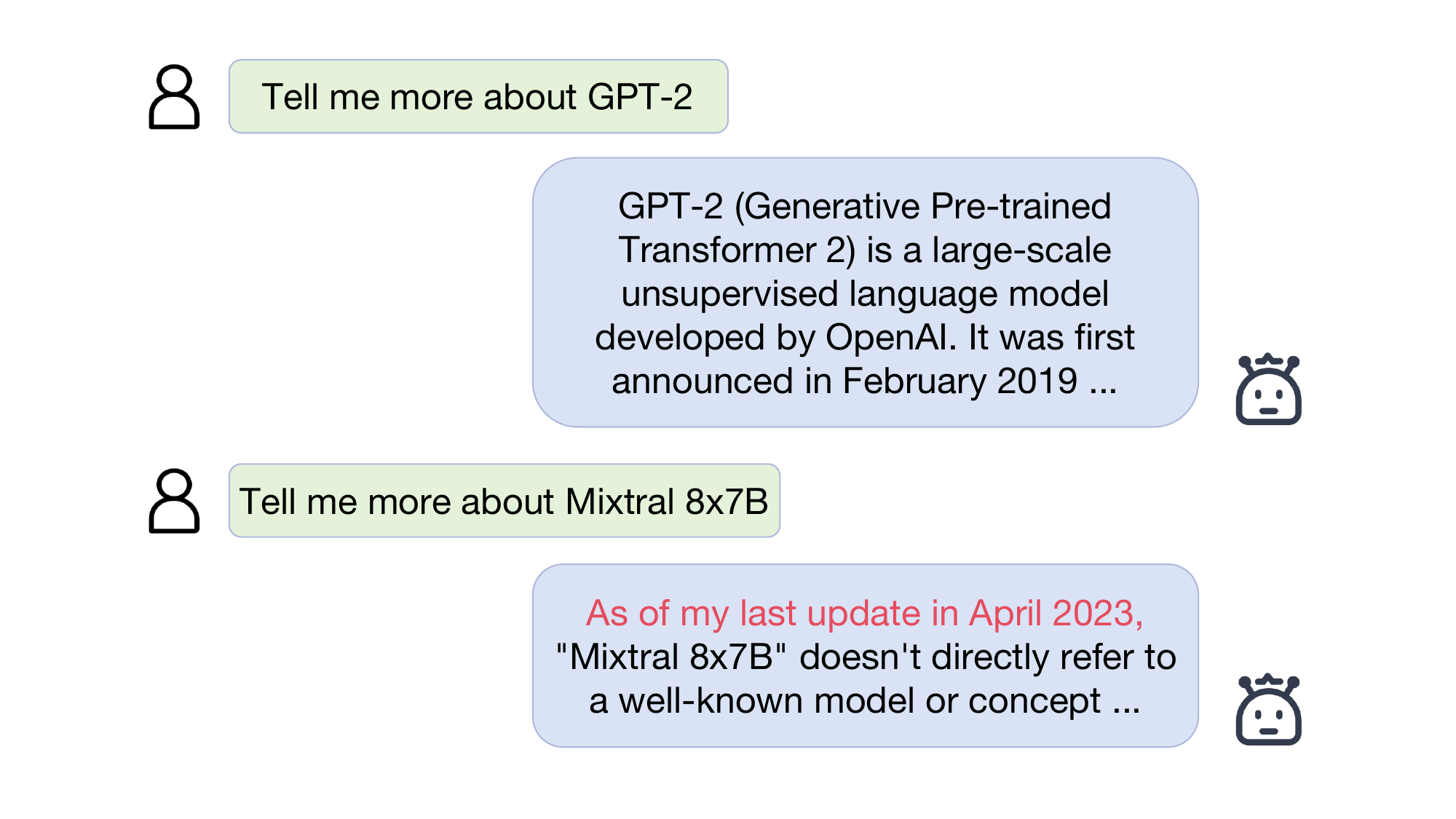}
\vspace{-3.7ex}
\caption{An example illustration, ChatGPT-4 is unable to provide accurate information about events that occurred after April 2023.}
\label{fig:exampleoutdate}
\vspace{-3.7ex}
\end{figure}

However, a significant challenge for LLMs is their restricted adaptability 
to the latest available data/information, which restraints them to
generate responses about recent events thus leading to information gaps. 
This entails hallucination, a phenomenon when LLM tries to generate plausible but incorrect answers about unknown facts~\cite{2023_hallucinate}. An example in this regard is shown in Figure~\ref{fig:exampleoutdate}, where ChatGPT-4 is unable to correctly answer a question about Mistral 8x7B, a model recently released in Dec 2023. Addressing issues such as outdated training data ~\cite{zhang2023large}, hallucination~\cite{zhang2023sirens}, and factual inaccuracies in LLMs~\cite{wang2023survey} is not only costly but also vulnerable to risks like model collapse~\cite{shumailov2023curse} and catastrophic forgetting~\cite{luo2023empirical}. Adapting these models to specific domains/tasks further intensifies these challenges~\cite{ling2023domain}.

It is important to make LLMs efficient lifelong learners. 
Recently,
there have been multiple different attempts to propose lifelong learning methods for knowledge updating~\cite{wu2023evakellm, zhang2024comprehensive} and skill acquisition~\cite{ling2023domain, zhang2023multitask, lewis2021retrievalaugmented}, a comprehensive overview of these existing strategies is provided in Appendix (Table \ref{tab:lllproblem_def}). However, existing approaches pose the following limitations: 
(i) These methods rely on sentences curated from fact 
triplets as the model's inputs, which is not practically feasible, 
as it is hard to organize all available information 
as structured units, e.g., a set of triplets; 
(ii) The majority of the existing approaches are vulnerable to catastrophic forgetting;
(iii) These approaches either focus on "open-book" or 
"closed-book" settings (Section \ref{sec:seeandremember}), with 
none of them providing an in-depth analysis of both approaches at 
the same time.
This calls the need for practical/easily adaptable data 
curation methodologies and accordingly better modeling 
strategies for the life-long learning of LLMs.

To address these challenges, in this paper, we propose \underline{\textbf{M}}ixture-\underline{\textbf{o}}f-Experts augmented Low \underline{\textbf{R}}ank \underline{\textbf{A}}daptation for \underline{\textbf{L}}ifelong learning (MoRAL). 
\OurMODEL{} simply relies on question-answer pairs for life-long learning.
Our key observation is: this architecture simultaneously exploits the multi-task modeling capability of the MoE structure and the parameter-efficient features of LoRA to achieve efficient lifelong learning. 
In order to test the effectiveness of \OurMODEL{} for different LLMs, 
we also introduce an evaluation benchmark, i.e., \underline{\textbf{L}}ife-\underline{\textbf{L}}ong 
\underline{\textbf{L}}earning of \underline{\textbf{LL}}Ms (5L-bench), encompassing a newly proposed dataset (question-answer pairs directly captured from the unstructured text rather than fact triplets) and novel evaluation 
metrics for performance comparison. We summarize the major 
contributions of this paper as follows:
\begin{enumerate}
\itemsep0em 
    \item We propose \textbf{MoRAL}, an effective strategy that combines the benefits of MoE along with LoRA as an effective and efficient strategy for lifelong learning of LLMs.
    \item We introduce a new evaluation benchmark, i.e.,  \textbf{5L-bench}, tailored to evaluating the lifelong learning abilities of LLMs using casual question-answer pairs from unstructured text 
    rather than fact triplets.
    \item We perform a rigorous evaluation of MoRAL under both "open-book" and "closed-book" settings. We delve into the interplay of these two methodologies, looking for insights, respective strengths, and limitations of MoRAL.
\end{enumerate}


\eat{\warn{why we set a new bench: A notable challenge identified is the diversity in data formats and evaluation criteria employed across methods, such as factual triplets for knowledge editing~\cite{decao2021editing,mitchell2022fast}, supervised Input-Output pairs~\cite{codealpaca,yue2023mammoth} for fine-tuning, and information chunks for Retrieval-Augmented Generation (RAG)~\cite{lewis2021retrievalaugmented}. This variety complicates dataset preparation and reuse. Also, in practice, we usually use a combination of methods to adapt LLMs to new domains and tasks~\cite{wang2023knowledgetuning,cui2023chatlaw}, when it is difficult to make evaluations with traditional evaluation pipelines that are isolated from each other. To fill in the gap, we introduce a novel benchmark}}

\eat{Specifically, unlike previous works that solely assess and compare the impact of model training on knowledge updating~\cite{wu2023evakellm,zhang2024comprehensive} and skill acquisition~\cite{ling2023domain,zhang2023multitask,lewis2021retrievalaugmented}, we consider retrieval augmentation~\cite{xu2023retrieval} as a crucial aspect of models' lifelong learning.  Our evaluation introduces two distinct pipelines: an open-book setup and a closed-book setup. This approach allows us to extend our analysis beyond simply evaluating the accuracy of model outputs. We place a significant emphasis on the models' abilities to summarize and filter information from contextual windows, providing a more comprehensive assessment of their lifelong learning performance. We test three open-source models of different scales using two efficient parameter fine-tuning schemes, LoRA and MoE-LoRA. We explore the interplay between the model's ability to learn from its context and the fine-tuning of the model's parameters.}

\eat{\asif{A significant challenge for LLMs is their restricted adaptability to the latest available data/information, which limits the abilities of LLMs to 
generate responses about recent events thus leading to information gaps. \eat{This entails hallucination, a phenomenon when LLM tries to 
generate plausible but incorrect answers about unknown facts.}
This is illustrated in Figure~\ref{fig:exampleoutdate}, where 
ChatGPT 4 is unable to provide correct information 
for a Mistral 8x7B model released in Dec 2023.} 
Addressing issues such as outdated training data~\cite{zhang2023large,yin2023alcuna}, 
hallucination~\cite{zhang2023sirens}, 
and factual inaccuracies in LLMs~\cite{wang2023survey} is not only 
costly but also~\asif{vulnurable to}~\eat{fraught with}risks like model collapse~\cite{shumailov2023curse} and catastrophic forgetting~\cite{luo2023empirical}. Adapting these models to specific domains~\asif{further} intensifies these challenges~\cite{ling2023domain}.
\asif{In order to address these challenges, in this research, we employ life-long learning to update 
the LLMs with new data. For this, we first clarify the concept of lifelong learning inspired by human's 
overall learning process. We then introduce an automated evaluation pipeline, 5L-Bench, to analyze the effectiveness and efficiency of lifelong learning methods in LLMs.}}

\eat{Therefore, it is important to make LLMs efficient lifelong learners.}

\eat{Lifelong learning, initially conceptualized by~\citet{thrun1995lifelong}, refers to a paradigm where a model leverages its previously acquired knowledge to enhance subsequent learning~\cite{thrun1995lifelong}. The primary features of lifelong learning include knowledge transfer and reuse, adaptation to new environments, and overcoming catastrophic forgetting~\cite{osti_1902727,new2022lifelong}.
\eat{With the advent of LLMs, the distinction between knowledge and skills is increasingly ambiguous. 
The definition of lifelong learning for these models still lacks clarity. So we first review existing strategies for efficient lifelong learning in LLMs. Drawing on the concept of lifelong learning in humans, values, skills, and knowledge~\cite{chong2009values} are of equal importance in the research on LLMs. Prior researches typically employ AI alignment to ensure that these models possess values aligned with humans~\cite{wang2023aligning,burns2023weaktostrong,ouyang2022training}.
During the pre-training phase, a broad array of datasets with diverse distributions were incorporated to equip the models with extensive knowledge~\cite{touvron2023llama,penedo2023refinedweb}. Furthermore, these models were continuously trained and evaluated on specific tasks to refine their skills~\cite{chen2021evaluating,zhong2017seq2sql,collins2023evaluating}. In our work, we emphasize how LLMs can efficiently acquire new knowledge and learn new tasks.}
In Figure \ref{fig:lifelongLLMdef}, we summarize the existing methods for lifelong learning in LLMs, which can be primarily categorized into two types. We conceptualize them as \asif{"remember it" and "see it"}~\eat{"see it" and "remember it"} (Figure \ref{fig:openclosedef}). The former represents adapting the model to different domain knowledge and application scenarios by modifying model parameters~\cite{wu2023bloomberggpt,poerner2020inexpensive,cui2023chatlaw}, akin to a closed-book exam where the model itself serves as the knowledge base~\cite{petroni2019language}. 
The latter signifies leveraging the model's language understanding and reasoning abilities by inserting the necessary knowledge or task examples into the model's context window~\cite{gao2024retrievalaugmented,parnami2022learning}, similar to an open-book exam where the model utilizes external knowledge. 
}
\eat{In Section \ref{section: related work}, we provide a comprehensive summary of the relevant works and characteristics for each specific method.}

\eat{\begin{figure}[b]
\centering
\includegraphics[width=0.35\textwidth]{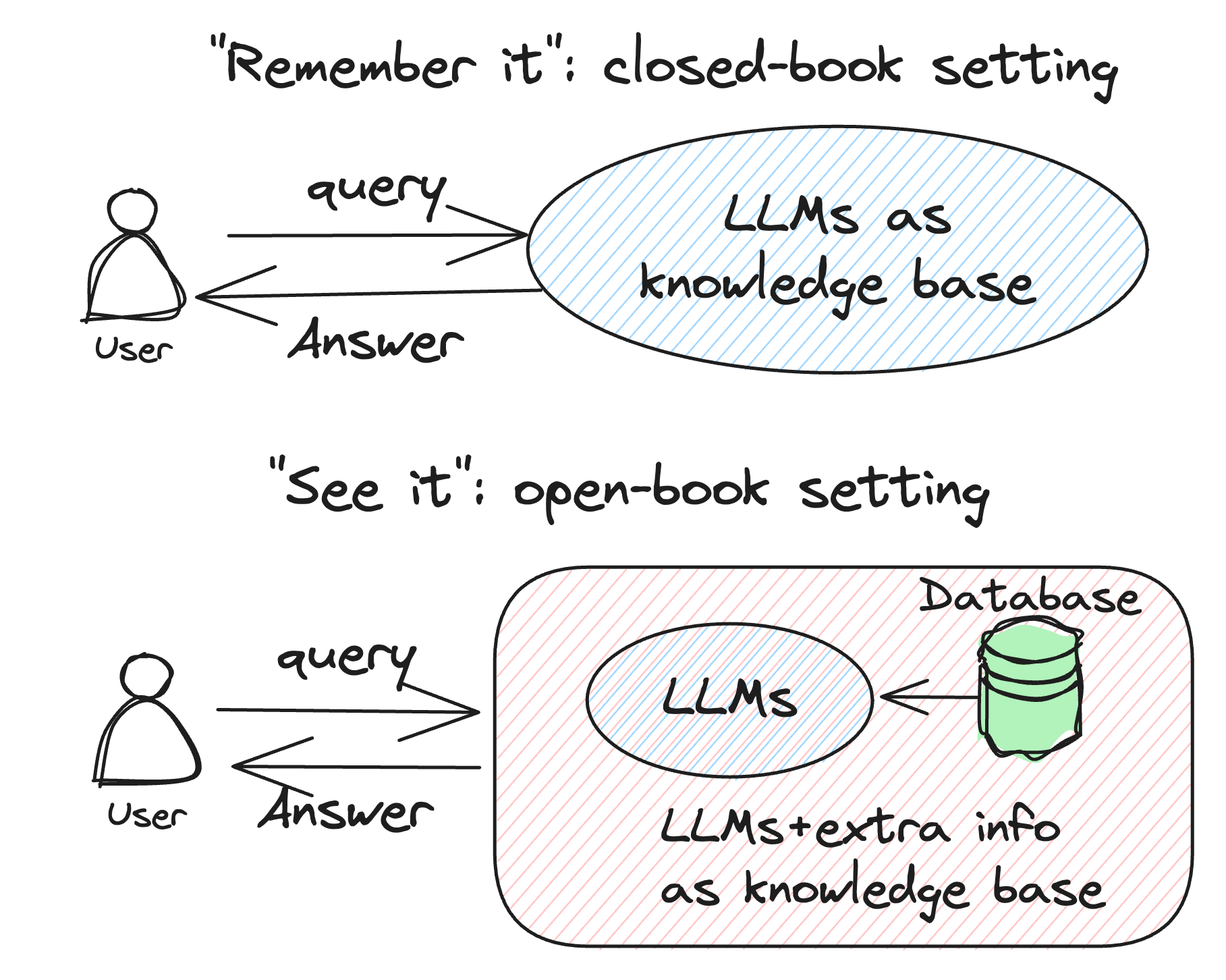}
\caption{Comparative illustration of closed-book vs. open-book settings in querying LLMs\footnote{\warn{\ref{sec:seeandremember} have detailed introduction about it}}}
\label{fig:openclosedef}
\vspace{-3.7ex}
\end{figure}
However, it remains unclear when to use various methods and their respective effectiveness~\cite{tu2022prompttuning,Nabwani2023}. This ambiguity arises due to two main reasons: Firstly, the evaluation of different methods' effectiveness depends on diverse downstream tasks. For instance, knowledge editing methods are often assessed based on the success rate of modifying information triples~\cite{wang2023easyedit,tran-etal-2022-improving}, while systems like retrieval augmented generation (RAG) adopt more flexible metrics such as faithfulness and answer relevance~\cite{es2023ragas}. Secondly, the interaction between different methods can influence their outcomes. For example, continued pre-training typically enhances model performance in the Supervised Fine-tuning (SFT) phase~\cite{kumar2022fine}. 

In this paper, we introduce the \textbf{5L-Bench} (Benchmark of \textbf{L}ife \textbf{L}ong \textbf{L}earning \textbf{L}arge \textbf{L}anguage Models) to evaluate whether the LLM is an effective lifelong learner.
\warn{Unlike previous works that solely assess and compare the impact of model training on knowledge updating~\cite{wu2023evakellm,zhang2024comprehensive} and skill acquisition~\cite{ling2023domain,zhang2023multitask,lewis2021retrievalaugmented}, we consider retrieval augmentation~\cite{xu2023retrieval} as a crucial aspect of models' lifelong learning. 
Our research aims to design an evaluation standard that supports both open-book and closed-book scenarios. We assess the impact of existing efficient fine-tuning methods (e.g., LoRA~\cite{hu2021lora} and MoE-LoRA~\cite{DBLP:journals/corr/abs-2310-18339}) on the models' performance and their capacity to leverage external data for answering questions. This lays the foundation for exploring and evaluating efficient lifelong learning methods for models, as well as the integration of model training with Retrieval-Augmented Generation (RAG) techniques.}\fixme{I think, this text needs to be coherent..!}}

\eat{
\asif{In this research, we plan to}
\eat{Therefore, it is important to} make LLMs efficient lifelong learners. Lifelong learning, initially conceptualized by~\citet{thrun1995lifelong}, refers to a paradigm where a model leverages its previously acquired knowledge to enhance subsequent learning~\cite{thrun1995lifelong}. The primary features of lifelong learning include knowledge transfer and reuse, adaptation to new environments, and overcoming catastrophic forgetting~\cite{osti_1902727,new2022lifelong}. With the advent of LLMs, the distinction between knowledge and skills is increasingly ambiguous. The definition of lifelong learning for these models still lacks clarity. So we first review existing strategies for efficient lifelong learning in LLMs. Drawing on the concept of lifelong learning in humans, values, skills, and knowledge~\cite{chong2009values} are of equal importance in the research on LLMs. Prior researches typically employ AI alignment to ensure that these models possess values aligned with humans~\cite{wang2023aligning,burns2023weaktostrong,ouyang2022training}. During the pre-training phase, a broad array of datasets with diverse distributions were incorporated to equip the models with extensive knowledge~\cite{touvron2023llama,penedo2023refinedweb}. Furthermore, these models were continuously trained and evaluated on specific tasks to refine their skills~\cite{chen2021evaluating,zhong2017seq2sql,collins2023evaluating}.In our work, we emphasize how LLMs can efficiently acquire new knowledge and learn new tasks.
In Figure \ref{fig:lifelongLLMdef}, we summarize the existing methods of lifelong learning in LLMs, which can be primarily categorized into two types. We conceptualize them as "see it" and "remember it" (Figure \ref{fig:openclosedef}). The former represents adapting the model to different domain knowledge and application scenarios by modifying model parameters~\cite{wu2023bloomberggpt,poerner2020inexpensive,cui2023chatlaw}, akin to a closed-book exam where the model itself serves as the knowledge base~\cite{petroni2019language}. The latter signifies leveraging the model's language understanding and reasoning abilities by inserting the necessary knowledge or task examples into the model's context window~\cite{gao2024retrievalaugmented,parnami2022learning}, similar to an open-book exam where the model utilizes external knowledge. In Section \ref{section: related work}, we provide a comprehensive summary of the relevant works and characteristics for each specific method.
\begin{figure*}[t]
\centering
\includegraphics[width=0.70\textwidth]{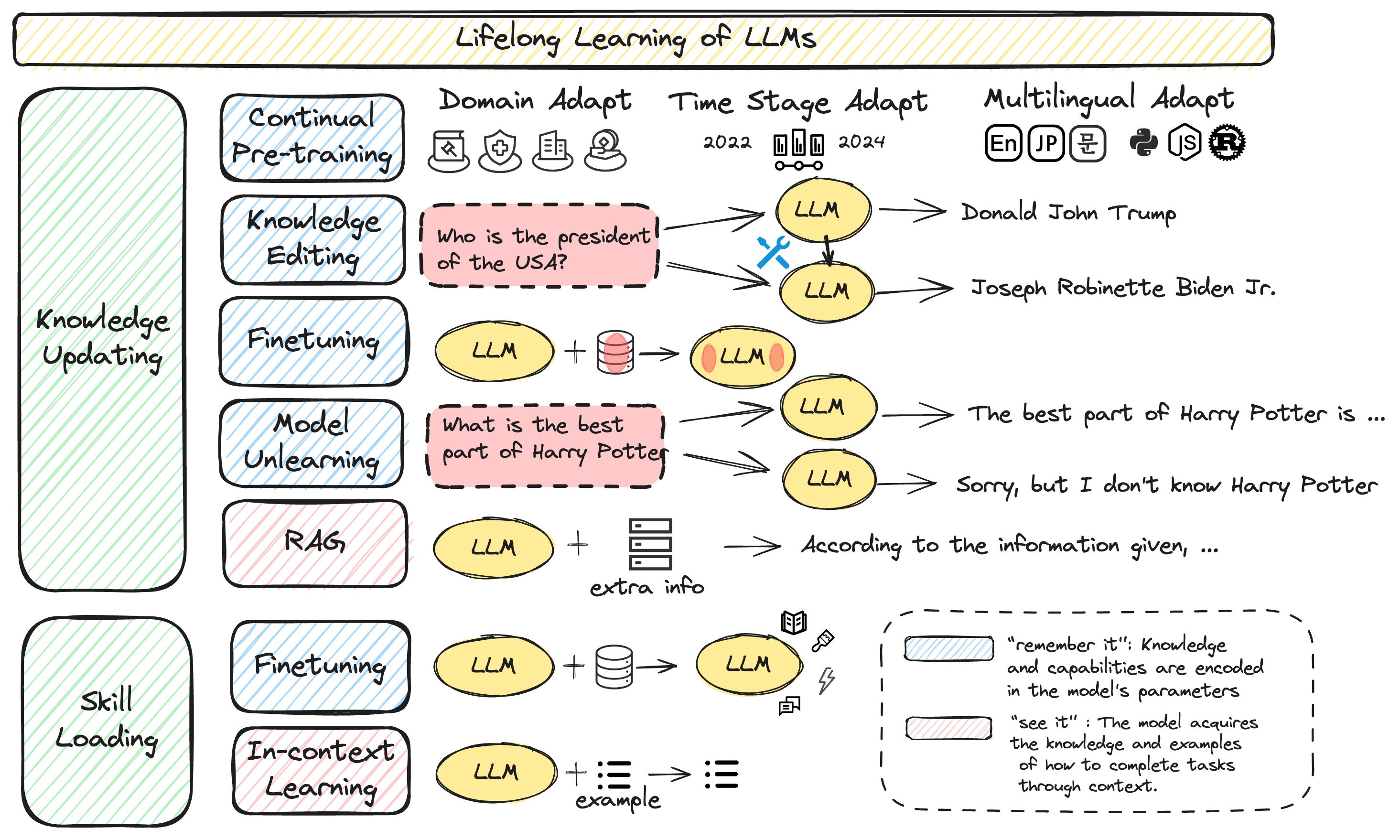}
\caption{Overview and examples of lifelong learning strategies for large language models.}
\label{fig:lifelongLLMdef}
\vspace{-2.7ex}
\end{figure*}
However, it remains unclear when to use various methods and their respective effectiveness~\cite{tu2022prompttuning,Nabwani2023}. This ambiguity arises due to two main reasons: Firstly, the evaluation of different methods' effectiveness depends on diverse downstream tasks. For instance, knowledge editing methods are often assessed based on the success rate of modifying information triples~\cite{wang2023easyedit,tran-etal-2022-improving}, while systems like retrieval augmented generation (RAG) adopt more flexible metrics such as faithfulness and answer relevance~\cite{es2023ragas}. Secondly, the interaction between different methods can influence their outcomes. For example, continued pre-training typically enhances model performance in the Supervised Fine-tuning (SFT) phase~\cite{kumar2022fine}. In this paper, we introduce the \textbf{5L-Bench} (Benchmark of \textbf{L}ife \textbf{L}ong \textbf{L}earning \textbf{L}arge \textbf{L}anguage Models) to evaluate whether the LLM is an effective lifelong learner.
\eat{\asif{A significant challenge for LLMs is their restricted adaptability to the latest available data/information, which limits the abilities of LLMs to 
generate responses about recent events thus leading to information gaps. \eat{This entails hallucination, a phenomenon when LLM tries to 
generate plausible but incorrect answers about unknown facts.}
This is illustrated in Figure~\ref{fig:exampleoutdate}, where 
ChatGPT 4 is unable to provide correct information 
for a Mistral 8x7B model released in Dec 2023.} 
Addressing issues such as outdated training data~\cite{zhang2023large,yin2023alcuna}, 
hallucination~\cite{zhang2023sirens}, 
and factual inaccuracies in LLMs~\cite{wang2023survey} is not only 
costly but also~\asif{vulnurable to}~\eat{fraught with}risks like model collapse~\cite{shumailov2023curse} and catastrophic forgetting~\cite{luo2023empirical}. Adapting these models to specific domains~\asif{further} intensifies these challenges~\cite{ling2023domain}.
\asif{In order to address these challenges, in this research, we employ life-long learning to update 
the LLMs with new data. For this, we first clarify the concept of lifelong learning inspired by human's 
overall learning process. We then introduce an automated evaluation pipeline, 5L-Bench, to analyze the effectiveness and efficiency of lifelong learning methods in LLMs.}}
}
\section{Related Works}
\label{section: related work}
\vspace{-0.7ex}

\noindent {\textbf{Continual Learning.}}
Continual learning (CL) aims~\eat{to deal with non-stationarity in training data. Its goal is}to learn new skills and knowledge without forgetting previous knowledge, also known as catastrophic forgetting~\cite{Kirkpatrick_2017,kaushik2021understanding}. \citet{maltoni2019continuous} delineated three principal strategies in CL: architectural~\cite{rusu2022progressive,lomonaco2017core50}, regularization~\cite{zenke2017continual}, and rehearsal~\cite{hayes2019memory}. 
They also conducted a thorough analysis of these strategies in sequentially 
learning incremental tasks. \citet{lesort2019continual,wang2023comprehensive} 
provided a comprehensive summary of lifelong learning from the perspective of 
autonomous agents. They emphasized that agents must adopt continuous methodologies for adaptation~\cite{sprechmann2018memorybased}, catastrophic forgetting, data distribution shifts~\cite{gepperth2016bio}. Also, in our experiments, we not only focus on the model's ability to learn new domain knowledge but also avoid catastrophic forgetting.

\noindent {\textbf{Lifelong Training of LLMs.}}
Continual learning offers a practical solution for adapting to novel data distributions~\cite{gururangan2020dont, xiong2023effective}. However, this approach is vulnerable to overfitting. {To mitigate this,~\citet{chen2023lifelong} introduced the Lifelong-MoE, an extensible architecture to allow pre-training over diverse data distributions}. Other than that, fine-tuning 
pre-trained foundation models also serve as an effective strategy for downstream task adaptation~\cite{zhou2023comprehensive, raffel2023exploring}. Amongst them, the parameter efficient variants include LoRA~\cite{hu2021lora}, Prompt Tuning~\cite{lester2021power}, etc. These methods optimize task-specific objectives by fine-tuning only a small set of parameters~\cite{peft,hu2023llm,yu2023melo,ling2023domain}. Motivated by these, in this paper we combine the multi-tasking modeling capability of the MoE structure and the parameter-efficient features of LoRA for an efficient lifelong training method.

\noindent {\textbf{Model Editing.}}
Model editing methods are used to make targeted, cost-effective fixes to edit the information contained in the LLMs~\cite{hartvigsen2023aging}. Existing model editing solutions support targeted operations, i.e., knowledge insertion, modification, and erasure~\cite{hase2023does, wen2023chathome, wang2023knowledge}. These methods may be categorized into: 
(i) meta-learning methods, which use external networks to predict gradients, e.g., MEND~\cite{mitchell2022fast} and 
(ii) locate-then-edit methods, which directly identify and update the target parameters, e.g., ROME~\cite{meng2023locating}, MEMIT~\cite{meng2023massediting}, etc., refer to~\citet{zhang2024comprehensive} for a recent survey. In-context learning methods resort to external knowledge~\cite{zheng2023edit}, and memory-based information retrieval~\cite{lewis2021retrievalaugmented, gao2024retrievalaugmented} to directly edit the model's knowledge~\cite{ovadia2024finetuning, pawelczyk2023incontext}. 

Key limitations of the existing methods is their reliance on 
factual triplets data, which creates challenges in data 
preparation and fully evaluating the effectiveness of the model performance~\cite{wu2023evakellm}, some challenges 
that are addressed by~\OurMODEL{}. 

\section{Preliminaries}
\noindent {\bf Notations:}
In this paper, we use $x$ to represent the input and $y$ as the output of the MoRAL architecture.
For the 5L-evaluation benchmark, we use 
$q$ to represent a query,
$C$ to represent the context.
$C_r$ represents the context fragments relevant to the query $q$,
$R_o$ represents the open-book response,  
$R_c$ represents the close-book response, 
and $G_t$ is the ground truth.

\subsection{"Open/Closed" book and Cross setting}
\label{sec:seeandremember}
"Open-book" and "closed-book" are two different strategies for querying LLMs. The major differences between these strategies are as follows:

\noindent \textbf{(a) Open-book.}
This strategy assumes that LLMs may refer to external data sources for inference. The external data sources may include 
but are not limited to databases, knowledge graphs, unstructured text, examples, etc.

\noindent \textbf{(b) Closed-book.} This strategy treats LLM as a data storage bucket that 
answers solely based on the knowledge gained during model training~\cite{alkhamissi2022review}.

\noindent \textbf{(c) Cross-Setting.}
The two settings ("open-book" and "close-book") are
interconnected. For this, we establish a criteria to 
investigate how enhancements in the model's closed-book 
capabilities simultaneously influence open-book setting. 
Likewise, there are some metrics equally important for 
both scenarios, e.g., fluency of the response.
To quantify this, we use "cross-setting" that evaluates 
all responses equally across different settings and 
computes the average scores.

\begin{figure}[t]
    \centering   
    \includegraphics[width=0.99\linewidth]{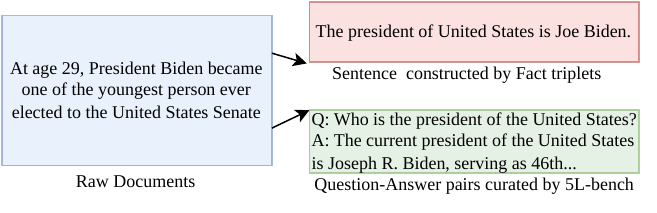}
    \vspace{-3.7ex}
    \caption{Example illustration of difference between the input data for 
    conventional approaches and \OurMODEL{}. }
    \label{fig:fact_Ex}
    \vspace{-3.7ex}
\end{figure}

\subsection{Fact Triples vs Question-Answer Pairs}
\label{sec:fact-triplet}
We illustrate the key differences between the input data format used by the
conventional approaches and~\OurMODEL{} in Figure~\ref{fig:fact_Ex}. For 
illustration, we want to update the knowledge of the model from 
\{"The president of United states is \emph{Donald Trump}"\} to "\emph{Joe Biden}".
The conventional methods will extract the relevant information triplet 
(president, Joe Biden, United States) from the raw documents, which will be later used to
formulate a sentence. Whereas our method (5L-bench) reformulates this 
information as question-answer pairs. We argue the latter approach is a more feasible and practical solution, as it is not possible to 
convert all available information as a set of triplets, leading to information loss.

\eat{In practice, these two settings can complement and interact with each other, so it 
is crucial to investigate the impact of "remembering" on proficiency in the 
open-book setting. This exploration can be paralleled to the concept of "practice 
makes perfect"~\cite{theophilides2000study,shimanovich2014testing}, examining 
how repetitive exposure and recall (in the stage of training) might enhance the 
model's ability to effectively utilize external resources during open-book tasks.}

\section{MoRAL for Lifelong LLMs}
As we mentioned, we aim to develop a lifelong learning method to keep LLMs up to date with the latest available knowledge and information. Unlike previous works relying on sentences directly curated using fact triplets, we use casual question-answer pairs directly captured from the unstructured text as the input (explained in Section~\ref{sec:fact-triplet}). For the lifelong learning strategy, 
we aim to combine the multi-task learning abilities of the MoE with the fine-tuning abilities of LoRA 
for effective learning. Specifically, we propose MoRAL (i.e., \underline{\textbf{M}}ixture-\underline{\textbf{o}}f-Experts augmented Low \underline{\textbf{R}}ank \underline{\textbf{A}}daptation for \underline{\textbf{L}}ifelong learning). 
MoRAL uses a divide-and-conquer strategy. It incorporates the benefits of 
using multiple experts along multiple different low-rank intrinsic knowledge dimensions with the hope of performing the end task in a performance-enhanced 
fashion.

The underlying motivation is that within the foundation LLMs, the 
knowledge/information resides along multiple different intrinsic/salient 
dimension, similar to subspaces~\cite{ali2019, hu2021lora}, and we may have multiple 
different localized/specialized experts to learn and/or override the prior information/knowledge contained by the LLM. 
We summarize the workflow of \OurMODEL{} as follows:
\begin{figure}[t]
\centering
\includegraphics[width=0.45\textwidth]{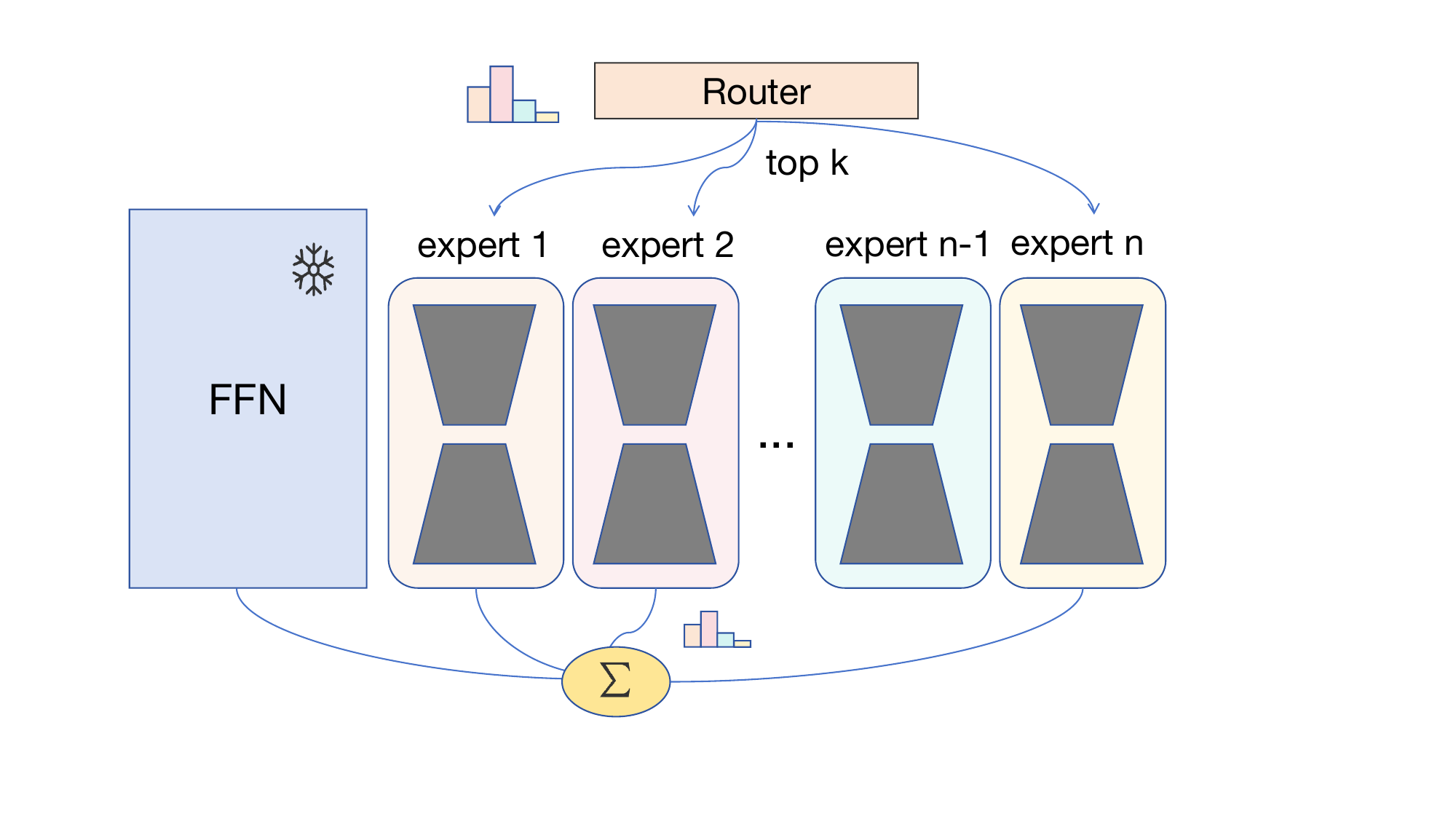}
\vspace{-3.7ex}
\caption{MoRAL architecture for life-long learning of LLMs. 
We use $n$ experts. FFN in the figure represents Feed-Foward Network.}
\vspace{-2.7ex}
\label{fig:Moelora}
\end{figure}

\begin{itemize}
\itemsep0em 
    \item Introduce low-rank matrices to decompose the weight matrices corresponding to the pre-trained LLMs.
    \item Use these low-rank matrices \eat{lightweight adapters} as experts 
    to be used on top of the pre-trained model.
    \item Allow conditional computation over multiple experts using a gating mechanism, also known as a router network.
\end{itemize}

\noindent For MoRAL, we configured eight LoRA expert modules, 
adopting a top-k routing strategy analogous to~\citet{jiang2024mixtral}. Figure~\ref{fig:Moelora} presents 
an illustration of the MoRAL structure. The computational steps 
for the router network and inference stages are explained as 
follows:

\paragraph{(a) Router Network.}
Assume there are $n$ localized experts, we use 
router network to compute the proportional score contribution of each expert.
The router network is defined as:
\begin{equation}
    \vspace{-0.7ex}
    G(x)_i = \text{softmax}(W_g^T x)
    \vspace{-0.7ex}
\end{equation}
where $W_g \in \mathbf{R}^{d_m \times n}$ represents the trainable 
weights of the router network with $d_m$ as the 
input dimension and $n$ as the number of experts.

\paragraph{(b) MoRAL Output.} The final output of the \OurMODEL{} 
architecture is computed as:
\begin{equation}
    \vspace{-1.1ex}
    y = \sum_{i=1}^{n} s_i \cdot E_i(x)
    \vspace{-0.7ex}
\end{equation}
where, $s_i = G(x)_i$ is the gating score for the $i^{th}$ expert, and $E_i(x)$ is the output from the expert for the input $x$.

\eat{Note, the end-goal of combining the MoE and LoRA architectures 
for life-long 
learning is to leverages the multi-task learning capabilities of 
MoE and the parameter-efficient fine-tuning benefits of LoRA. 
We hope with the hope to mitigate issues such as: ability to 
deal with imbalanced data and catastrophic forgetting.}


\section{5L-Bench (Evaluation Benchmark)}
\label{sec:5lbench}
For the performance evaluation of~\OurMODEL{}, we propose a 
new benchmark (i.e., 5L-bench). It encompasses:
(1) A new curated dataset namely: Arxiv, to test the ability of~\OurMODEL{} to adapt to new data domains. (2) A pre-existing dataset, i.e., HotpotQA~\cite{yang2018hotpotqa} used to test the ability of \OurMODEL{} to restrain knowledge by not allowing the model to forget old knowledge. (3) Newly proposed evaluation metrics to rigorously evaluate 
    the performance of \OurMODEL{} under open-book, closed-book 
    and cross-settings.

\subsection{Arxiv Data Curation}
\label{sec:Arxiv-data}
\begin{figure}[t]
\centering
\includegraphics[width=1.05\linewidth]{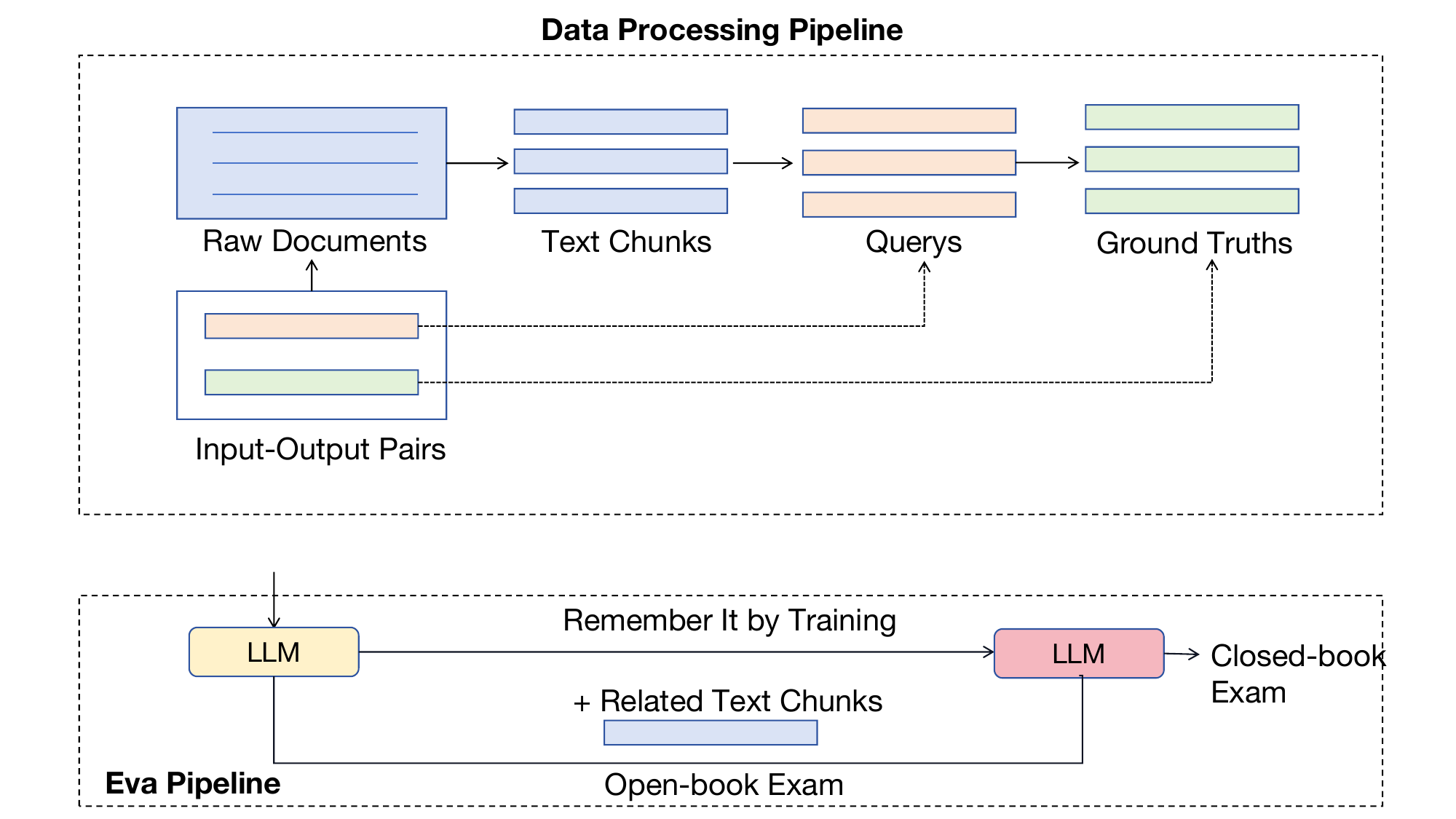}
  \caption{Overview of the 5L-Bench data curation and evaluation pipeline.}
  \label{fig:5Lframework}
\vspace{-2.7ex}
\end{figure}

Our data curation pipeline is shown in the upper-half of Figure~\ref{fig:5Lframework}.
It aims to curate a set of question-answer pairs from unstructured 
text, and it is explained as follows.

Firstly, we acquire unlabeled raw documents from Arxiv and 
split them into information chunks $C$. 
Then, we employ GPT-3.5-turbo-16k to generate the corresponding questions $q$ for each chunk $c \in C$~\cite{li2023camel}. 
Following this, we use GPT-4 to generate the ground truth $G_t$, 
creating standard answers based on the questions 
and their associated information.

 Data leakage is a key challenge when it comes to evaluating LLMs 
on vast datasets~\cite{li2024open}. To prevent the model from having 
prior exposure to the data we intend to use for model training, 
we use the latest papers, i.e., from December 2023 Arxiv, as our data source.
To ensure precise document segmentation and facilitate data generation in a format conducive to 
our analysis, we utilize the method of recursively splitting by character\footnote{\url{https://python.langchain.com/docs/modules/data_connection/document_transformers/recursive_text_splitter}}. Additionally, we leverage prompts detailed in Appendix~\ref{sec:appendix_prompt} to guide the model to output data in the desired format.

The data curation process generates a quintuple dataset denoted as: 
\{$q$: query, $C$: context, $C_r$: retrieved contexts, $R_o$: open-book response, $R_c$: closed-book response, $G_t$: ground truth\}. Note that in our configuration, each query $q$ is uniquely paired with a context $c \in C$. The retrieved context set $C_r$ encompasses fragments from the context $c$ that exhibit relevance to the query. 
This relevance is computed by the cosine similarity between the embeddings of $q$ and each context in $C$, exceeding a predefined threshold $\theta$. 
\begin{equation}
\label{eq:thr_measure}
\small
C_r = \{ c \in C \mid \cos(\text{EMB}(q), \text{EMB}(c)) > \theta \},
\end{equation}
where $\text{EMB}(\cdot)$ denotes the text embeddings, and $\cos(\xx,\yy)$ denotes the cosine similarity between vectors $\xx$ and $\yy$. 
Note, this is a widely adopted method for semantic-based 
information retrieval~\cite{reimers-2020-multilingual-sentence-bert, reimers-2019-sentence-bert}.

\subsection{Evaluation Metrics}
\label{sec:Eval}
5L-bench uses different evaluation metrics to test \OurMODEL{} 
under open-book, closed-book and cross settings. Details are 
as follows:
\paragraph{(a) Open-book Settings.}
For the open-book settings, we aim to explore the ability of 
LLMs to utilize external information within the context window~\cite{xu2024retrieval}. 
A key concern is to investigate if the model leverages the 
additional knowledge for reasoning or merely for replication. 
For this, we design an evaluation criterion sub-divided 
into four distinct scenarios: 
\begin{itemize}
\itemsep0em 
    \item \textbf{Context Faithfulness (Faith).} When the model is provided with "golden context", that is, $C_r = \{c\}$, our criterion focuses more on the consistency between LLM and external information to see whether\eat{specifically whether} the final answer conflicts with the given context. 
    \item \textbf{Irrelevant Context Filtering (Filter).} When the model uses external information that encompasses $c$ along with other unrelated contexts, i.e., $c \in C_r$ \text{and} $|C_r| > 1$, our criterion prioritizes how well LLM avoids answering irrelevant or context.
    \item \textbf{Refusal Rate (RR).} When LLM is presented with external data entirely unrelated to the question, i.e., $c \notin C_r$, we assess the ability of LLM to refuse to answer the query. \eat{beyond their ability. by refusing questions beyond its capabilities}
    \item For the cases with $C_r = \emptyset$, we will employ the same metrics used in the closed-book setting.
\end{itemize}

\paragraph{(b) Closed-book Settings.}
For the closed-book settings, we focus on learning objectives, 
i.e., we test the new knowledge or capabilities for LLMs. 
For this, we use Recall Accuracy (RA) as the evaluation
metric~\cite{derczynski-2016-complementarity,es2023ragas}. 
The computation of RA is illustrated in Appendix~\ref{Appendix:eval}.

\paragraph{(c) Cross Settings.}
For cross-settings, we evaluate: (i) the compliance of model's 
response with the given query, i.e., how well the model answers 
a given question also referred to as \textbf{Query Relevance 
(QR)} by~\citet{es2023ragas}; and (ii) the model's linguistic 
modeling capabilities, assessed via the \textbf{fluency (FL)} 
of the model's responses. Details on "QR" and "FL" are provided in 
the Appendix~\ref{Appendix:eval}.

\section{Experimentation}

\subsection{Experimental Setup}
\noindent {\bf Datasets.} For performance evaluation, we use newly 
curated data (Arxiv) and an open-source data HotpotQA~\cite{yang2018hotpotqa}.  
The Arxiv dataset comprises seven domains with a diverse range of 
topics from mathematics to artificial intelligence. 
We use this dataset as the target data for learning new knowledge.
\eat{Similar to~\citet{Guyon1997ASL}, w}
We split this data into 80\%  and 20\% for training and test sets, respectively.
The HotpotQA dataset encompasses 1,500 rows. We use 
this data as the hold-out data for testing knowledge-retaining ability.
The statistics of the data are given in Appendix~\ref{append:data} (Table~\ref{tab:eval_data}).

\noindent {\bf Experimental Settings.}
In order to train \OurMODEL{}, we utilize the 
Adam optimizer~\cite{2014_adam} with a learning rate of 0.0001. 
The batch size is set to 16, and the model is trained for 2 epochs.
As shown in Figure \ref{fig:Moelora}, we apply MoRAL to the frozen FFN layers.
We use {the} number of experts (n = 8), and top $k$ = 2. For Equation~\ref{eq:thr_measure}, 
we use $\theta$ = 0.87. All experiments are performed using Pytorch and Nvidia A100 80G GPU.

\noindent {\bf Large Language Models.}
For experimental evaluation, we use multiple different 
open-source and closed-source LLMs. Specifically, we use basic large language models TinyLlama-1.1B~\cite{zhang2024tinyllama}, Phi-2-2.7B\footnote{\url{https://huggingface.co/microsoft/phi-2}}, Llama2-7B~\cite{touvron2023llama}, and state-of-the-art (SOTA) closed-source LLMs including 
GPT-3.5-turbo-16k\footnote{\url{https://platform.openai.com/docs/models/gpt-3-5-turbo}}, 
Gemini-pro~\cite{gemini}, and 
Claude-2.1~\cite{claude}. 
Details about these 
models are in the Appendix~\ref{appendix:LLMs}.

\noindent  {\bf Baselines.}
\eat{In our exploration of open-source models, w}
We use multiple parameter-efficient fine-tuning approaches 
as baselines, namely: 
(a) LoRA~\cite{hu2021lora}, 
(b) IA3~\cite{liu2022fewshot}, and 
(c) LLaMA-Adapter~\cite{zhang2023llamaadapter}.
It is notable that our model is not comparable with the existing 
knowledge-editing and life-long learning baselines, e.g., MELO by~\citet{yu2023melo}, MEND by~\citet{mitchell2022fast} etc., 
as these models rely on factual triplets as the model input, 
which makes them different from our work. Details about 
the baseline approaches are in the Appendix~\ref{appendix:baseline}.

\noindent {\bf Evaluation Workflow.}
Our evaluation is structured around three primary settings, 
i.e., open-book, closed-book and cross configurations (see Section \ref{sec:seeandremember}). 
In the closed-book setting, the model generates responses 
solely based on its internal knowledge following the given 
instructions.
For the open-book setting, we employ the bge-large-en-v1.5 model~\cite{bge_embedding} for embedding generation, coupled 
with \textit{chroma}\footnote{\url{https://www.trychroma.com/}} as the vector database to store the embeddings of text blocks 
{$c \in C$}. 
It allows us to identify text blocks with cosine similarity scores against the query exceeding a predefined threshold \(\theta\).
{These blocks, denoted as \(C_r\)}, are then inserted into the 
model's context window, guiding the model to evaluate the 
relevance between questions and answers. This process enables 
the model to autonomously refine, filter, and if necessary 
decline to respond based on the context's relevancy and the instructions provided. 
The response output of the model is finally assessed to 
measure the disparity between the generated answer and 
the ground truth using the evaluation metrics explained 
in Section~\ref{sec:Eval}.
To mitigate the risk of bias arising from the use of 
a single model~\cite{zeng2023evaluating,hada2023large} in our 
evaluation, we employ GPT-4-1106-preview and GLM-4
\footnote{\url{https://zhipuai.cn/devday}} as evaluators. 
We use the average scores of these evaluators as the final 
assessment metric.

\eat{This bifurcated evaluation approach facilitates a 
comprehensive understanding of the model's performance in 
both knowledge-integrated and standalone response scenarios.}


\eat{We designed experiments to explore whether using 
low-resource and efficient methodologies could enable 
these "smaller" models to achieve comparable performance to proprietary larger closed-source LLMs on new distributions 
of datasets. Additionally, by selecting models of varying 
sizes and training on datasets of different scales, we aimed 
to investigate their characteristics when faced with new 
datasets requiring learning and adaptation.}
\subsection{Experimental Results}

\begin{table*}[!ht]
    \centering
    \small
    \resizebox{0.72\textwidth}{!}{
    \begin{tabular}{lllllcll}
    \toprule
        \multirow{2}{*}{\textbf{Models}} & \multicolumn{4}{c}{\textbf{Open-book}}& \textbf{Closed-book}& \multicolumn{2}{c}{\textbf{Cross-setting}}\\ 
        \cmidrule{2-8}
        & Faith.↑ & Filter.↑ & RR.↑ & RA↑ & RA↑ & QR↑  & FL.↑\\ 
        \midrule
        TinyLlama-1.1B-Chat & 0.65 & 0.40 & 0.24 & 0.86 & 0.60 & 0.82  & \textbf{0.95}\\ 
        TinyLlama-1.1B-Chat+IA3& 0.54& 0.38& 0.25& 0.82& 0.64& 0.82& 0.90\\ 
        TinyLlama-1.1B-Chat+LLaMA-Adapter& 0.66& 0.36& 0.29& 0.74& 0.67& 0.89& 0.91\\ 
        TinyLlama-1.1B-Chat+LoRA& \textbf{0.69}& 0.43 & \textbf{0.32}& 0.89 & \textbf{0.82}& 0.85  & 0.90\\ 
        TinyLlama-1.1B-Chat+MoRAL& 0.63 & \textbf{0.58}& 0.28 & \textbf{0.91} & 0.77 & \textbf{0.90}& 0.93
\\ 
        \hline
        Phi-2-2.7B& 0.54 & 0.31 & 0.33 & 0.73 & 0.41 & \textbf{0.88}& \textbf{0.89}\\ 
        
        Phi-2-2.7B+IA3& 0.55& 0.28& 0.28& 0.62& 0.40& 0.80& 0.83\\ 
        Phi-2-2.7B+LLaMA-Adapter& 0.59& 0.30& 0.35& 0.69& 0.48& 0.84& 0.85\\
        
        Phi-2-2.7B+LoRA& 0.47 & 0.35 & \textbf{0.39}& 0.77 & \textbf{0.66}& 0.80& 0.84
\\ 
        Phi-2-2.7B+MoRAL& \textbf{0.59}& \textbf{0.46}& 0.37 & \textbf{0.82}& 0.63 & 0.86  & 0.88
\\ 
        \hline
        Llama-2-7B-chat-hf& 0.62 & 0.54 & 0.40& 0.82 & 0.47 & 0.80& \textbf{0.92}\\ 
        Llama-2-7B-chat-hf+IA3& 0.67& 0.50& 0.43& 0.77& 0.50& 0.75& 0.86\\ 
        Llama-2-7B-chat-hf+LLaMA-Adapter& 0.61& 0.52& 0.37& 0.67& 0.54& 0.77& 0.89\\
        Llama-2-7B-chat-hf+LoRA& 0.65 & 0.50& 0.48 & 0.83 & 0.72 & 0.83  & 0.87
\\ 
        Llama-2-7B-chat-hf+MoRAL& \textbf{0.71}& \textbf{0.61}& \textbf{0.51}& \textbf{0.90}& \textbf{0.79}& \textbf{0.92}& 0.90\\ 
        \hline
        GPT-3.5-turbo-16k & 0.80 & 0.64 & 0.75 & 0.92 & 0.73 & 0.92  & \textbf{0.97}\\ 
        Gemini-pro\footnote{\url{https://ai.google.dev/models/gemini}} & 0.83 & 0.82 & 0.79 & 0.83 & \textbf{0.80}& 0.95  & 0.95
\\ 
        Claude-2 & \textbf{0.92}& \textbf{0.87}& \textbf{0.90}& \textbf{0.96} & 0.77 & \textbf{0.98}& 0.95\\
        \bottomrule
  \end{tabular}}
  \vspace{-1.7ex}
    \caption{\OurMODEL{} performance comparison against different LLMs using Arxiv data. All evaluation metrics range from 0 to 1, with higher scores indicating better performance. We boldface the best-performing scores.}
    \label{tab:results_newdomain}
    \vspace{-12pt}
\end{table*}

\paragraph{LLMs Learn Fast in "Open-book".}
Table \ref{tab:results_newdomain} shows the results of
\OurMODEL{} on the Arxiv dataset, compared against the baseline models.
We use the notation "+(strategy)" to specify the corresponding 
fine-tuning strategy employed by the LLM. 

For the open-source LLMs without any fine-tuning, we observe that exposing the large model solely to the relevant context within the contextual window for inference significantly enhances its performance. This is evident by an increased Recall Accuracy (RA) score for TinyLlama-1.1B, i.e., 0.86 in open-book settings, compared to 0.6 in the closed-book setting. Likewise, the performance of Phi-2-2.7B and Llama-2-7B improves 
significantly, i.e., 0.73 and 0.82 in open-book compared to 0.41 and 0.47 in closed-book respectively.

A similar trend is observed for the closed-source LLMs, with GPT-3.5-turbo, Gemini-pro, and Claude-2 improving the "RA" by 26.0\%, 3.7\%, and 24.6\% for the open-book settings compared to the closed-book settings. Despite fine-tuning, closed-source LLMs continue to outperform open-source smaller models, particularly in terms of metrics "Faith", "Filter", and "RR". This superiority could stem from the large models' effective human-alignment strategies~\cite{ouyang2022training}, which enhance their contextual understanding and adherence to instructions. This suggests that the real disparity between small open-source and proprietary large models may lie deeper in their ability to model the comprehension of language and tasks~\cite{brown2020language, sun2024survey} rather than their capacity to generate responses aligned with standard answers. 

Overall results showcase the immense potential for integrating dynamic information retrieval methods in LLMs' context for enhanced performance. These results strongly correlate with earlier studies by~\citet{balaguer2024rag} and ~\citet{zheng2023edit} that emphasize the significance of retrieval-augmented generation 
for large models~\cite{gao2024retrievalaugmented}. 

\paragraph{\OurMODEL{} vs Baselines.} 
Comparing the results of~\OurMODEL{} against the baseline 
models, we observe for "RA" metric,~\OurMODEL{} consistently 
outperforms the baseline models in the open-book settings 
with very few exceptions.
For instance, compared to the pre-trained models,~\OurMODEL{} improves 
the "RA" score for TinyLlama-1.1B, Phi-2-2.7B and Llama-2-7B 
by 5.81\%, 12.32\% and 9.75\% respectively in open-book settings.
{Whereas, for the closed-book settings, TinyLlama-1.1B and Phi-2-2.7B 
models fine-tuned using LoRA exhibit slightly better or comparable 
"RA" scores compared to~\OurMODEL{}, except for Llama-2-7B which 
performs best when fine-tuned using~\OurMODEL{}.}

{Comparing the results for other open-book metrics, i.e., "Faith", "Filter", and "RR", we observe: 
(i) For the metric "Faith", LLMs trained using~\OurMODEL{} results 
a higher score except for TinyLlama-1.1B where LoRA performs slightly better;
(ii) for the metric "Filter"~\OurMODEL{} consistently outperforms all 
baseline models by a significant margin;
(iii) for "RR" results of \OurMODEL{} are comparable with the
baseline models. These results strongly portray the immense 
potential of \OurMODEL{} when employed in the open-book settings.}

{For the cross-settings, we observe that~\OurMODEL{} 
results in higher "QR" with very relatively low distortion 
in the fluency (FL) compared to baselines. This decrease in the "FL" after instruction fine-tuning is likely due to the prevalence of scientific descriptions and mathematical formulas in our data, which reduced the model's general language modeling capability~\cite{belle2023exploring}. Notably, the decline in fluency was less pronounced for the models fine-tuned using~\OurMODEL{}.}

\begin{figure}[t]
\centering
\includegraphics[width=0.33\textwidth]{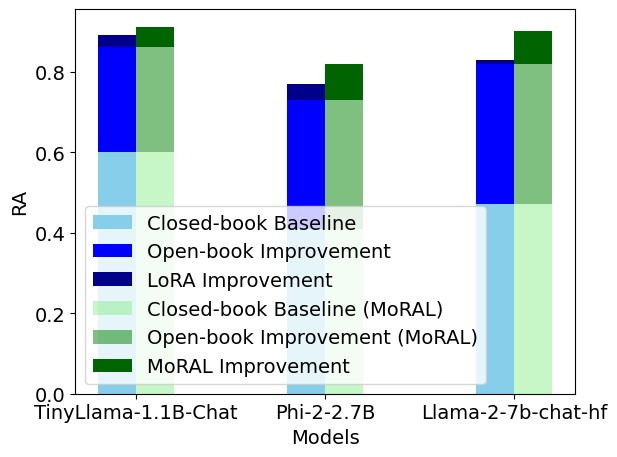}
\vspace{-11pt}
\caption{Performance comparison of~\OurMODEL{} vs LoRA
for large models with varying number of model parameters, best viewed in colors. 
These results are computed using the Arxiv dataset.}
\label{fig:LoRAMoeloraresult}
\vspace{-3.7ex}
\end{figure}

{We also observe, as the scale of the model increases, 
the model's capabilities in filtering and analyzing 
information from context increases significantly. 
This is also evident by a relative higher scores for the metrics: 
"Faith", "Filter", and "RR" for the large models fine-tuned 
using \OurMODEL{} vs models with relatively lower parameters.}
This is also illustrated in Figure \ref{fig:LoRAMoeloraresult},
where the dark green region shows that~\OurMODEL{} yields a 
higher improvement in the "RA" score for Phi-2-2.7B and Llama-2-7B compared to 
that of TinyLlama-1.1B. It also shows that relative improvement 
in performance for \OurMODEL{} is higher compared to LoRA as a baseline.

To summarize, these results show \OurMODEL{} 
presents a promising direction for effective and efficient 
learning of LLMs. This ascertains our hypothesis that
the multi-tasking ability of a mixture of experts when 
coupled with LoRA significantly augments the contextual 
learning ability of the end model, also shown previously on multiple
different tasks~\cite{zoph2022stmoe, xue2024openmoe}.

\subsection{Further Discussions}
In this section, we perform an in-depth analysis in 
attempts to understand the life-long learning of 
LLMs from multiple different perspectives. 

\begin{table*}[ht]
    \centering
    \small
    \resizebox{0.72\textwidth}{!}{
    \begin{tabular}{lccccccc}
    \toprule
        \multirow{2}{*}{\textbf{Models}} & \multicolumn{4}{c}{\textbf{Open-book\eat{(model+RAG)}} \textbf{} \textbf{}}& \textbf{Closed-book}& \multicolumn{2}{c}{\textbf{Cross-setting}}\\
        \cmidrule{2-8}
        & Faith.↑ & Filter.↑ & RR.↑ & RA↑ & RA↑ & QR↑  & FL.↑\\ 
        \midrule
        TinyLlama-1.1B-Chat-v1.0& 0.67 & 0.41 & 0.20& \textbf{0.89} & \textbf{0.72} & \textbf{0.90}& 0.95 \\ 
        TinyLlama-1.1B-Chat+IA3& 0.60& 0.33& \textbf{0.21}& 0.73& 0.64& 0.80& 0.89\\ 
        TinyLlama-1.1B-Chat+LLaMA-Adapter& 0.61& 0.39& 0.20& 0.75& 0.64& 0.81& 0.88\\ 

        TinyLlama-1.1B-Chat+LoRA& \textbf{0.68}& 0.38 & 0.17 & 0.87 & 0.65 & 0.83 & 0.91\\ 
        TinyLlama-1.1B-Chat+MoRAL& 0.67 & \textbf{0.43}& \textbf{0.21}& \textbf{0.89}& 0.70& 0.87& \textbf{0.96}\\ 
        \hline
        Phi-2-2.7B& \textbf{0.55}& \textbf{0.33}& \textbf{0.42}& \textbf{0.76}& \textbf{0.68}& \textbf{0.93}& 0.92 \\ 
        
        Phi-2-2.7B+IA3& 0.46& 0.33& 0.40& 0.73& 0.60& 0.88& 0.90\\ 
        Phi-2-2.7B+LLaMA-Adapter& 0.49& 0.30& 0.23& 0.70& 0.55& 0.90& 0.89\\
        
        Phi-2-2.7B+LoRA& 0.49 & 0.27& 0.31& 0.71& 0.64& 0.80& 0.90\\ 
        Phi-2-2.7B+MoRAL& 0.53& 0.30& 0.37& 0.74& 0.65& 0.86& \textbf{0.93}\\ 
        \hline
        Llama-2-7B-chat-hf& \textbf{0.72}& \textbf{0.62}& \textbf{0.51}& \textbf{0.90}& \textbf{0.75}& \textbf{0.95}& \textbf{0.96}\\ 
        
        Llama-2-7B-chat-hf+IA3& 0.63& 0.50& 0.45& 0.86& 0.71& 0.88& 0.92\\ 
        Llama-2-7B-chat-hf+LLaMA-Adapter& 0.60& 0.48& 0.41& 0.80& 0.68& 0.89& 0.86\\
        
        Llama-2-7B-chat-hf+LoRA& 0.65& 0.47& 0.43& 0.85& 0.71& 0.89& 0.91\\ 
        Llama-2-7B-chat-hf+MoRAL& 0.69& 0.58& 0.48& 0.89& 0.71& 0.90& 0.95\\ 
        \bottomrule
  \end{tabular}}
  \vspace{-1.7ex}
    \caption{\OurMODEL{} performance for different LLMs using HotpotQA-fullwiki dataset. All evaluation metrics range from 0 to 1, with higher scores indicating better performance. We boldface the best performing scores.}
    \label{tab:results_reserve}
\vspace{-12pt}
\end{table*}

\noindent {\bf More Data or More Parameters?} 
We first aim to answer the question: "In terms of data 
and model parameters, what is required to make the end-model 
a better lifelong learner?"

Surprisingly, we observe among the pre-trained LLMs (w/o fine-tuning), 
TinyLlama-1.1B with only 1.1B parameters shows the best "RA" 
performance compared to other baselines, i.e., RA is 0.60 and 
0.86 in closed-book and open-book settings respectively 
(Table~\ref{tab:results_newdomain}).
It showcases the potential of "small" language models 
trained on vast datasets. This finding is also aligned with
a recent work, where a relatively small model, i.e., 
MiniCPM~\cite{minicpm2024}
with only 2B parameters is able to outperform 13B models on 
UltraEval\footnote{\url{https://github.com/OpenBMB/UltraEval}}. 

However, a model with fewer parameters, i.e., TinyLlama-1.1B, 
exhibits significantly lower scores for the open-book 
metrics ("Faith", "Filter", and "RR"), 
compared to larger counterparts, with "RR" showing the most 
substantial disparity—TinyLlama-1.1B's baseline score is 0.24,
compared to 0.4 for Llama-2-7B. 
This shows that larger models are more adept at 
declining questions beyond their comprehension 
scope~\eat{within context windows}. It also 
speaks of larger models' enhanced in-context learning 
ability~\cite{DBLP:journals/corr/abs-2303-03846}, 
enabling them to better filter and summarize 
information. 
\eat{Additionally, the degree of alignment between the 
model and humans can influence their fidelity to context 
and perception of capability boundaries. However, our 
current work does not explore whether larger models 
possess greater alignment potential, a topic we aim 
to address in future research.}

\eat{
Experimental evaluation reveals the potential of the MoE structure in addressing complex learning scenarios, demonstrating its superiority in multitasking and mitigating 
catastrophic forgetting.}

\noindent {\bf Learning New without Forgetting Old.}
\begin{figure}[t]
\centering
\includegraphics[width=0.48\textwidth]{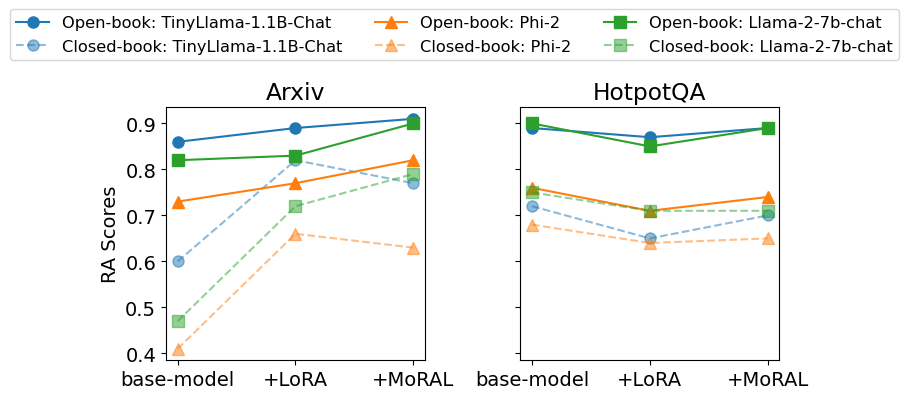}
\vspace{-3.7ex}
\caption{\OurMODEL{} performance comparisons for "RA". The left half of the 
Figure reports results on Arxiv data from Table~\ref{tab:results_newdomain}.
The right half of the Figure reports results on HotpotQA data from Table~\ref{tab:results_reserve}.}
\label{fig:inoutdomain}
\vspace{-3.7ex}
\end{figure}
Adjusting parameters in large models risks catastrophic 
forgetting, a critical challenge in lifelong learning that
emphasizes the need for adapting to new domains/tasks 
without losing prior 
knowledge~\cite{new2022lifelong,luo2023empirical}. 
To test the knowledge retention ability 
of~\OurMODEL{} compared against baselines, we 
use the HotpotQA dataset as a holdout test set 
to re-evaluate models fine-tuned using the Arxiv dataset.
Corresponding results in Table~\ref{tab:results_reserve}
show that the baseline models yield a lower score for the
metrics: "Faith", "Filter", "RR", etc.~\OurMODEL{} on the 
other hand, results in minimal loss for these metrics while 
at the same time it shows proficiency in instruction 
compliance and language fluency (FL).  

Correlating the results for Table~\ref{tab:results_newdomain} and Table~\ref{tab:results_reserve}, we observe that although baseline 
fine-tuning approaches significantly boost "RA" scores 
for new target domains, however, they yield a decline in "RA" 
score for the holdout tests.~\OurMODEL{} on the other 
hand exhibits better resistance to catastrophic 
forgetting by exhibiting a relatively stable performance.
This is also illustrated in Figure \ref{fig:inoutdomain}.
The left half of the Figure shows~\OurMODEL{} augments
the knowledge retention ability compared to baseline while 
learning new knowledge.
The right half of the Figure shows that for the HotpotQA
data, where the LoRA baseline yields lower "RA" scores, whereas
\OurMODEL{} fights back to uplift the "RA" score. 
Overall knowledge retention ability of~\OurMODEL{} is more
pronounced for the open-book scenarios.

Note, for these experiments, we observe higher initial scores 
for the HotpotQA dataset for the pre-trained base models, possibly 
because: 
(i) The HotpotQA dataset and its Wikipedia sources were 
part of LLMs' training data; 
(ii) The training datasets include knowledge from 2018, aligning 
with the view that models are biased to answer questions from 
this period due to temporal information encoded in their 
parameters~\cite{nylund2023time}.

\eat{\warn{
\noindent {\bf Enhance the “Deeper” Capabilities of LLMs.} In our experiments, we found that enhancing the model's ability for contextual summarization and filtration proves to be more challenging than merely improving its alignment with standard answers. This is why there is a lot of work devoted to improving the complex logical reasoning of models~\cite{huang2023reasoning}. Although existing work has shown that pre-trained foundational models largely determine the upper limit of a model's performance, the cost of pre-training from scratch is enormous~\cite{kaplan2020scaling}. How to enable continuous improvement of LLMs is a proposition worth exploring. We will continue to investigate efficient methods to enhance the deep capabilities of models in our future work.}}

\vspace{-2.1ex}
\section{Conclusions}
\vspace{-2.1ex}
In this paper, we make following contributions:
(i) we propose~\OurMODEL{} for efficient and effective 
life-long learning of LLMs; 
(ii) we propose an evaluation benchmark (5L-bench) to 
evaluate the performance of \OurMODEL{} compared 
against the baseline models. 
In the future, we plan to explore larger-scale models 
and more efficient hybrid structures, such as Mixture 
of Vectors (MoV) by~\citet{zadouri2023pushing}.
\section{Limitations}
\paragraph{Surface Learning vs. Deep Understanding.}
Although, this paper shows that fine-tuned models are able to 
achieve significant improvements in both open-book 
and closed-book settings.
Still, this work did not evaluate if the models only superficially 
learn to produce answers that conform more closely to 
standard responses without truly familiarizing themselves 
with the knowledge and concepts within the training data.

\paragraph{Reliability of LLMs as Evaluators.} 
In our work, both GPT-4 and GLM-4 are employed as evaluators 
to mitigate the bias that may arise from relying on a single 
model for assessment\cite{hada2023large}. 
Although large language models are extensively being used in evaluating 
language tasks, demonstrating higher consistency compared 
to human evaluators. Yet, employing more robust models to assess 
less advanced counterparts essentially guides the models 
towards alignment with the evaluator's characteristics~\cite{lin2023llmeval}.
This alignment could potentially limit the models' capacity to align 
with human understanding, thereby constraining their 
performance upper bounds.

\bibliography{custom}
\bibliographystyle{acl_natbib}

\clearpage
\appendix
\section{Existing Challenges in Lifelong Learning}
\label{sec:appendix}

Lifelong learning, initially conceptualized by~\citet{thrun1995lifelong}, refers to a paradigm where a model leverages its previously acquired knowledge to enhance subsequent learning~\cite{thrun1995lifelong}. The primary features of lifelong learning include knowledge transfer, adaptation to new environments, and overcoming catastrophic forgetting~\cite{osti_1902727,new2022lifelong}. With the advent of LLMs, the distinction between knowledge and skills is increasingly ambiguous. 
The definition of lifelong learning for these models still lacks clarity.\eat{In Figure \ref{fig:lifelongLLMdef} and Table \ref{tab:lllproblem_def},} We summarize the existing lifelong learning methods along with their corresponding evaluation metrics in Table~\ref{tab:lllproblem_def}.
Briefly, these existing approaches can be divided into two categories: one is to train the model to "remember" new knowledge and skills (closed-book); the other is to put additional information into the model's context window so that the model can "see it" and make a response (open-book). 

We observe a notable challenge for lifelong learning of LLMs is the diversity in data formats, such as factual triplets~\cite{decao2021editing, mitchell2022fast}, 
supervised input-output pairs~\cite{codealpaca,yue2023mammoth}, and information chunks ~\cite{lewis2021retrievalaugmented}. Such a vast diversity of data formats complicates data preparation and reuse.
There is a dire need to use simple data preparation strategies along with robust lifelong learning methods.
Also, in practice, we usually use a combination of methods to adapt LLMs to new domains and tasks~\cite{wang2023knowledgetuning,cui2023chatlaw} which makes it difficult to make evaluations with traditional evaluation pipelines that are isolated from each other. 

\begin{table*}[t]
\vspace{2.7ex}
\small
\centering
\resizebox{0.80\textwidth}{!}{
\begin{tabular}{p{3cm}|p{4cm}|p{8cm}}
\toprule
\textbf{Methodologies} & \textbf{Scenarios} & \textbf{Elements} \\ 
\midrule
Continual Pre-training & Out-Distribution Adaptation (for better downstream tasks performance) & PPL~\cite{jelinek1977perplexity}; Forget R.~\cite{Liu_2020}; MF1; Acc~\cite{gururangan2021demix,aghajanyan2021muppet}\\ \hline
\multirow{3}{*}{Knowledge Editing} & Knowledge Insertion & \multirow{3}{8cm}{Reliability; Generalization~\cite{zhang2024comprehensive}; Portability; Locality~\cite{yao2023editing}; Fluency\cite{meng2023locating}; Cross-lingual Evaluation (CKEE)~\cite{wu2023evakellm}} \\
\cline{2-2} 
                                   & Knowledge Modification & \\ 
\cline{2-2} 
                                   & Knowledge Erasure & \\ 
\hline
Fine-tuning & Downstream Tasks & F1 (political affiliation classification); ROUGE-L (news summarization)~\cite{Dhingra_2022}; BLEU (Machine Translation)~\cite{PapineniRWZ02};CLUES~\cite{mukherjee2021clues} \\ 
\hline
Model Unlearning & Knowledge Erasure & UnlearningSuccess (Generic tasks)~\cite{pawelczyk2023incontext}; Unlearning Harmfulness (Security and Alignment)~\cite{yao2023large} \\ 
\hline
RAG & Create,Read,Update and Delete (CRUD)~\cite{CRUDRAG} & ROUGE, BLEU, bertScore, RAGQuestEval~\cite{CRUDRAG}, Ragas~\cite{es2023ragas} \\ 
\hline
In-context Learning& Downstream Tasks & F1 (political affiliation classification); ROUGE-L (news summarization)~\cite{Dhingra_2022}; BLEU (Machine Translation)~\cite{PapineniRWZ02} \\ 
\bottomrule
\end{tabular}}
\vspace{-1.7ex}
\caption{Different methodologies and evaluation metrics for lifelong learning.}
\label{tab:lllproblem_def}
\vspace{-2.7ex}
\end{table*}

\section{Details of Experiments}
\subsection{Evaluation Metrics}
\label{Appendix:eval}
\eat{
\paragraph{(a) \textbf{TP}}(True Positives): Statements that are present in both the answer and the ground truth.
\paragraph{(b) \textbf{FP}}(False Positives): Statements present in the answer but not found in the ground truth.
\paragraph{(c) \textbf{FN}}(False Negatives): Relevant statements found in the ground truth but not present in the answer.}

\paragraph{(a) \textbf{QR}}(Query Relevance):
QR measures how well the response is aligned with the 
input query/question ($q$). For the computation of QR, 
we use the same settings as that of~\citet{es2023ragas}. 
We use context $c$ and response
$R$ in order to generate the question $Q(R,c)$. Later, we
compute the similarity between the generated question and query $q$. This score is computed as:
\begin{equation}
    \text{QR} = \frac{1}{n} \sum_{i=1}^{n}\text{sim}(q,Q(a,c))
\end{equation}
where sim is the cosine similarity of the corresponding embedding
vectors.

\paragraph{(b) \textbf{FL}}(Fluency): 
FL measures if the text generated is well-written and grammatical. \citet{fang2023chatgpt, wu2023chatgpt} have shown the 
remarkable capabilities of large language models in 
assessing sentence fluency and grammatical accuracy, 
highlighting their superiority over conventional 
approaches. In our experimental framework, we employ 
GPT-4 and GLM-4 for FL evaluation. The prompts utilized 
in our study are delineated in 
Table~\ref{table: eva prompt template FL}.
{The} final FL score is computed as:
\begin{equation}
 \text{FL} = \frac{1}{n} \sum_{i=1}^{n} \text{mean} (\text{GPT-4}, \text{GLM-4})
\end{equation}

\paragraph{(c) \textbf{RA}}(Recall Accuracy):
In order to compute RA, we first compute: \textbf{TP} (True Positives), 
\textbf{FP} (False Positives), and \textbf{FN} (False Positives). Then RA
is computed as:

\begin{equation}
\small
\text{RA}=\frac{F1 \times w_0 + \cos(\text{EMB}(a), \text{EMB}(G_t)) \times w_1}{w_0 + w_1}
\end{equation}
where the $F1$ score is computed as:
$\text{TP}/(\text{TP} + 0.5 \times (\text{FP} + \text{FN}))$.
In above equation, $\text{EMB}(y)$ and $\text{EMB}(G_t)$ represent the embeddings of the model's output and the ground truth.
The weights $w_0$ and $w_1$ are used to balance the F1 score 
and the cosine similarity of the embeddings.

\subsection{Large Language Models}
\label{appendix:LLMs}
\paragraph{(a) TinyLlama-1.1B-Chat-v1.0.}
TinyLlama-1.1B, a relatively smaller model compared to Llama, was pre-trained on 3 
trillion tokens~\cite{zhang2024tinyllama} and fine-tuned on the Ultrachat~\cite{UltraChat} dataset.

\paragraph{(b) Phi-2-2.7B.}
Phi-2-2.7B\footnote{\url{https://huggingface.co/microsoft/phi-2}}, a model with 2.7 billion parameters, was trained on a dataset comprising 1.4 trillion tokens, including a substantial 
number of textbooks~\cite{gunasekar2023textbooks}. 
\eat{and demonstrates outstanding reasoning and language understanding capabilities. 
This model showcases state-of-the-art performance among base language models with fewer than 13 billion parameters.}

\paragraph{(c) Llama2-7b-chat.}
Llama2-7b-chat is an open-source model pre-trained on 2.0 trillion tokens, fine-tuned on publicly available instruction datasets, as well as over one million new human-annotated examples~\cite{touvron2023llama}. 

\paragraph{(d) SOTA Closed-source LLMs.}
Among the closed-source LLMs, we compare~\OurMODEL{} against state-of-the-art (SOTA) models, including: GPT-3.5-turbo-16k~\cite{brown2020language}, Gemini-pro~\cite{gemini}, and Claude-2.1~\cite{claude}. These models are accessed via API calls.

\begin{table}[!ht]
\centering
\resizebox{0.48\textwidth}{!}{
\begin{tabular}{c|c}
\toprule 
\textbf{Symbols} & \textbf{Meaning} \\
\midrule 
$q$ & the query \\
$C$ & the context \\
$C_r$ & the context fragments relevant to the query $q$ \\
$R_o$ & the open-book response \\ 
$R_c$ & the close-book response \\
$G_t$ & ground truth \\
\bottomrule
\end{tabular}}
\vspace{-1.7ex}
\caption{Notations.}
\label{notation}
\end{table}

\subsection{Baselines}
\label{appendix:baseline}
\paragraph{(a) LoRA.}
LoRA uses a set of trainable rank decomposition matrices for the Transformer layers fine-tuning phase~\cite{hu2021lora}. In our case, we use LoRA adaptors for the attention layer, i.e., for the query ($q$) and key ($k$) matrices to enable efficient learning.

\paragraph{(b) IA3.} 
IA3 re-calibrates internal activations by suppressing and amplifying them, thus injecting adapters through the modulation of internal activations. These learned vectors are integrated into the attention and feed-forward modules of typical Transformer-based architectures~\cite{liu2022fewshot}. In our case, IA3 weights are added to the outputs of the key and value layers, as well as the input to the second feed-forward layer in each Transformer block.

\paragraph{(c) LLaMA-Adapter.} The Llama-Adapter is designed to adapt the Llama model for instruction following tasks. To avoid introducing noise into the tokens, the adapter employs zero-init attention~\cite{zhang2023llamaadapter}. Additionally, the adapter incorporates a learnable gating factor, also initialized to zero, which allows for the gradual introduction of information to the model during training.

\subsection{Data Statistics}
\label{append:data}
The statistics of the dataset is shown in Table~\ref{tab:eval_data}.
\begin{table}[h]
\small
\centering
\resizebox{0.5\linewidth}{!}{
\begin{tabular}{c|c}
\toprule
\textbf{Domain} & \textbf{Size}\\ \midrule
    Arxiv-Math & 1,518 rows \\ 
    Arxiv-Astro-ph & 1,811 rows \\ 
    Arxiv-Gr-qc & 1,749 rows \\ 
    Arxiv-Q-bio & 1,749 rows \\ 
    Arxiv-Q-fin & 2,513 rows \\ 
    Arxiv-Statistics & 2,208 rows \\ 
    Arxiv-EESS & 1,442 rows \\ 
    Arxiv-Ai & 2,001 rows \\ 
    \midrule
    HotpotQA-fullwiki & 1,500 rows \\ 
\bottomrule
\end{tabular}}
\vspace{-1.7ex}
\caption{Dataset distribution of different datasets, i.e., Arxiv and HotpotQA.}
\label{tab:eval_data}
\vspace{-12pt}
\end{table}

\subsection{Prompts}
\label{sec:appendix_prompt}
In this section, we present a detailed overview of the prompts used for data generation. Table \ref{table: prompt template 0} demonstrates the prompts designed for generating queries from various data sources. Following this, Table \ref{table: prompt template 1} describes the prompts used for generating ground truth data used for evaluating model performance. Table \ref{table: eva prompt template openbook} and Table \ref{table: eva prompt template closedbook} cover the methodologies for prompt generation in open-book and closed-book settings, respectively. 

\clearpage

\begin{table}[!ht]
\centering
\begin{tabular}{|m{0.95\linewidth}|}
\hline
\begin{verbatim}
You are a University Professor creating 
a test for advanced students. For each 
context, create a question that is 
specific to the context. Avoid creating 
generic or general questions.
Context: {context}
Question: a question about the context.
Format the output as *JSON* with the 
following key:
"question" \n\n
\end{verbatim}
\\
\hline
\end{tabular}
\vspace{-1.7ex}
\caption{The prompt template for generating the \texttt{query} from existing \texttt{context}.}
\label{table: prompt template 0}
\end{table}

\begin{table}[!ht]
\centering
\begin{tabular}{|m{0.95\linewidth}|}
\hline
\begin{verbatim}
You are a University Professor creating 
a test for advanced students. For each 
question and context, create a standard 
answer.
Context: {context} 
Question: {question}
    
Format the output as *JSON* with the f
ollowing keys:
“ground truth” \n\n
\end{verbatim}
\\
\hline
\end{tabular}
\vspace{-1.7ex}
\caption{The prompt template for generating the \texttt{ground truth}.}
\label{table: prompt template 1}
\end{table}

\begin{table}[!ht]
\centering
\begin{tabular}{|m{0.95\linewidth}|}
\hline
\begin{verbatim}
Please answer the question based only 
on the following context. If you cannot 
answer the question with the context, 
please respond with 'I don't know':
    ### CONTEXT
    {context}
    ### QUESTION
    Question: {question}
    \n
Please give your answer below:
\end{verbatim}
\\
\hline
\end{tabular}
\vspace{-1.7ex}
\caption{The prompt template for open-book setting.}
\label{table: eva prompt template openbook}
\end{table}

\begin{table}[!ht]
\centering
\begin{tabular}{|m{0.95\linewidth}|}
\hline
\begin{verbatim}
Please answer the following question.
    ### QUESTION
    Question: {question}
    \n
Give your answer below:
\end{verbatim}
\\
\hline
\end{tabular}
\vspace{-1.7ex}
\caption{The prompt template for closed-book setting.}
\label{table: eva prompt template closedbook}
\end{table}

\begin{table}[!ht]
\centering
\begin{tabular}{|m{0.95\linewidth}|}
\hline
\begin{verbatim}
Evaluate the fluency of a given piece 
of text on a scale from 0 to 1, where 
0 represents very poor fluency with 
numerous grammatical errors and awkward 
phrasing, and 1 represents excellent 
fluency with smooth, natural language 
and no grammatical mistakes. Consider 
aspects such as grammar, syntax, c
oherence, and the natural flow of 
ideas. Please provide a clear rating 
and a brief justification for your 
assessment, highlighting specific 
examples from the text that influenced 
your rating. Your evaluation should
be flexible enough to accommodate 
a variety of texts while maintaining
a focus on fluency and coherence.
\n
##Text
Text: {response}
\n
Give your score below:
\end{verbatim}
\\
\hline
\end{tabular}
\vspace{-1.7ex}
\caption{The prompt template for FL evaluation.}
\label{table: eva prompt template FL}
\end{table}

\end{document}